\documentclass{article}

\usepackage{arxiv}

\usepackage[utf8]{inputenc} 
\usepackage[T1]{fontenc}    
\usepackage{hyperref}       
\usepackage{url}            
\usepackage{booktabs}       
\usepackage{amsfonts}       
\usepackage{nicefrac}       
\usepackage{microtype}      
\usepackage{lipsum}
\usepackage{graphicx}
\graphicspath{ {./images/} }

\title{A Brief Discussion on KPI Development in Public Administration }

\author{
 Simona Fioretto \\
  Department of Electrical and Information Technology Engineering\\
  University of Naples Federico II\\
  Naples, Italy\\
  \texttt{simona.fioretto@unina.it} \\
  \And
 Elio Masciari \\
  Department of Electrical and Information Technology Engineering\\
  University of Naples Federico II\\
  Naples, Italy\\
  \texttt{elio.masciari@unina.it} \\
  \And
 Enea Vincenzo Napolitano \\
  Department of Electrical and Information Technology Engineering\\
  University of Naples Federico II\\
  Naples, Italy\\
  \texttt{eneavincenzo.napolitano@unina.it} \\
}

\begin{document}
\maketitle
\begin{abstract}
Efficient and effective service delivery in Public Administration (PA) relies on the development and utilization of key performance indicators (KPIs) for evaluating and measuring performance. This paper presents an innovative framework for KPI construction within performance evaluation systems, leveraging Random Forest algorithms and variable importance analysis. The proposed approach identifies key variables that significantly influence PA performance, offering valuable insights into the critical factors driving organizational success. By integrating variable importance analysis with expert consultation, relevant KPIs can be systematically developed, ensuring that improvement strategies address performance-critical areas. The framework incorporates continuous monitoring mechanisms and adaptive phases to refine KPIs in response to evolving administrative needs. This study aims to enhance PA performance through the application of machine learning techniques, fostering a more agile and results-driven approach to public administration.
\end{abstract}


\section{\uppercase{Introduction}}
\label{sec:introduction}
The success of an organisation depends on its ability to meet internal and external objectives. This involves the alignment of the mission and strategy of the organisation with the needs of its customers. In fact, once the needs of the customers are identified, they must be translated into organisational goals driving mission and strategy of the organization. Then, to evaluate the achievement of these goals, organisations require an objective measurement system. 
In fact, the ability of an organization of performing activities by pursuing efficiency and efficacy, is a measure of its performance results. Therefore, measuring performance is a complex and structured system. In fact, performance is a multifaceted phenomenon that requires integrated and simultaneous analysis of several indicators. Individual indicators often capture only a portion of the complexity of the organization, which instead is influenced by many variables. The identification of Key Performance Indicators (KPIs) aligned with the objectives of the organization, is crucial for assessing the performance in the organization  and identify areas for improvement \cite{banu2018measuring}. These indicators use quantitative metrics to summarise information about specific phenomena of interest to stakeholders \cite{jahangirian2017key}. Indeed, to evaluate whether a process adheres to policies, meets deadlines, or is able to respect a fixed budget, it may be necessary to use a combination of multiple indicators. 
In the context of Public Administration (PA), the implementation of a proper performance measurement system can be fundamental to assure high quality services to citizens. However, the definition of KPIs in the PA sector is not as simple as it can be in private companies.  In fact, the PAs significantly differ from the dynamics mechanism which are typical of the private sector. This is due to PAs characteristics. In fact, PAs differ from each other for the offered services, and for offices characteristics, such as the number of citizens served, the number of employees, and the level of office digitization \cite{kerzner2019using}. For instance, in justice sector or in education, only simple and measurable indicators are needed such as the required average time to resolve a legal case or graduation rates\cite{amato2023evolving}. However, these simple measures, which are called in the following macro-KPI, may not fully capture the quality level or fairness of the services provided. Additionally, PAs face challenges with bureaucracy and resistance to change. Administrative procedures can oppose to the adoption or modification of KPIs, even when they are no longer effective. The definition and interpretation of KPIs can also be heavily influenced by political environment, with changes in administration sometimes resulting in a complete restructuring of objectives and evaluation metrics.   
Accountability is another critical dimension in PA. Unlike the private sector, where accountability is primarily focused on financial results, in PAs the accountability is towards citizens. This is translated into the need of higher levels of transparency and communication, with resulting understandable KPIs by the public. One of the main issues in KPI definition, is the imprecise definition of objectives to which parameters should be aligned. If the objectives are unclear, the chosen parameters may not be relevant, resulting in the collection of meaningless data that do not provide insights into performance levels. In addition, markets and operating environments evolve quickly, and parameters previously defined may no longer be appropriate for the current necessities.

However, the use of digital technologies in the context of PA can properly support the objective definition and monitoring of KPIs based on data. By leveraging the information contained in operational data, it is possible to define objectively KPIs, guaranteeing a more suitable performance management system. In this paper we aim to introduce a framework, which starting from macro-KPI, leverages data for identifying the specific micro-KPI. Micro-KPIs investigate and measure the variables results leading to the results of the macro-KPIs.
In particular, can be leveraged the power of Machine Learning to select the most influential features, which can be used to properly define micro-KPIs which contribute to simpler macro-KPIs. 

The paper is structured as follows: Section \ref{th} introduces the theoretical concepts fundamental to our study. We discuss the nature and importance of KPIs and outline the characteristics of the chosen ML model. Section \ref{RW} reviews existing literature on the use of KPIs in PA and the construction of KPIs through machine learning techniques. Finally, in Section \ref{fp}, we propose an innovative framework for constructing KPIs based on the use of machine learning. This framework aims to enhance the accuracy and relevance of KPIs used in PA by utilising the potential of Random Forest to analyse and interpret large volumes of data. The approach demonstrated in this text provides practical and meaningful insights for the evaluation and optimisation of PA processes.

\section{\uppercase{Theoretical Background}}
\label{th}
This section provides the theoretical background required throughout the paper, offering basic concepts of Key Performance Indicators (KPIs) and Random Forest. 
The theoretical background is essential to understand the innovations proposed in our research framework, which will be outlined in the following sections.\\

\subsection{Key Performance Indicators}
Organizations continuously set goals in order to achieve better results in terms of efficiency and efficacy. These goals are both a translation of the mission and the strategy of the organization. They need to be objectively monitored, in order to understand the status of their achievement. In fact, by monitoring their activities, organisations can determine whether or not they have achieved their objectives \cite{dominguez2019taxonomy}. The evaluation of goals achievement can be done by defining objective metrics, known as KPI. KPIs are a collection of crucial measures, both financial and non-financial, that are utilised to convert objectives into tangible measures. In details, the authors in \cite{dominguez2019taxonomy} demonstrate that KPIs can provide organisations with reliable information to establish the basis for implementing growth strategies. KPIs can provide a way to see whether the strategic plan being adopted is working, serving as a tool to drive desired behaviours, and that their use can increase and improve operational efficiency, productivity and profitability. By establishing a set of KPIs, an organization can evaluate whether it has reached its goals \cite{velimirovic2011role}. The relationship between the success of the organization and KPIs is evident, as they are closely linked to  goals achievement.

\subsection{Importance factor for Random Forest}
The Random Forest (RF) method \cite{parmar2019review} is an ensemble learning technique for classification, regression, and other tasks. It constructs multiple decision trees during the training phase and outputs the mode of the classes (classification) or the mean of the predictions (regression) of the individual trees. The model's robustness is enhanced by its ability to withstand variation without significantly increasing bias, thanks to its natural ensemble.
One significant contribution of RF is its ability to assess the importance of variables, known as feature importance, in the predictive model. This is typically calculated in two ways \cite{strobl2008conditional}:
    \begin{enumerate}
        \item \textbf{Importance based on decreasing impurity}: is a method used to measure the importance of a variable in decision trees. It calculates how much the Gini index or entropy decreases due to the splits made on that variable. This method aggregates the total decrease in impurity attributable to each variable across all forest trees, normally weighted by the number of observations passing through those splits.
        \item \textbf{Importance of Allowed Variance}: This text evaluates the impact of a variable by mixing its values across observations in the test dataset. If there is a significant decrease in model performance after permutation, it indicates a high importance of the mixed variable. This is because its direct alteration deteriorates the model's ability to make accurate predictions.
    \end{enumerate}

These methods for evaluating variable importance are essential not only for optimizing RF models but also for providing insights into the characteristics that have the greatest impact on the target variable, thereby offering guidance for understanding and interpreting performance to define KPIs.

\section{\uppercase{Related Work}}
\label{RW}
This section reviews the literature and research related to the identification and development of Key Performance Indicators (KPIs) in Public Administration (PA) and the application of Machine Learning (ML) techniques to enhance these processes. The discussion is organized into two main subsections, corresponding to the primary focus areas of the research.

\subsection{The Identification of Key Performance Indicators in Public Administration}

The selection of appropriate KPIs requires consideration of several factors. Numerous studies have addressed KPI definition and selection in organizations. For instance, Parmenter \cite{parmenter2015key} proposes a framework that takes into account organizational characteristics, emphasizing the importance of aligning KPIs with Critical Success Factors (CSFs). Specifically for PAs, further assumptions must be made, linking KPIs to the perspectives of the Balanced Scorecard (BSC). Parmenter \cite{parmenter2012key} provides a comprehensive methodology tailored to PAs, considering their non-profit nature. Unlike private organizations focused on budget optimization, PAs prioritize measuring variables tied to service quality and resource efficiency. KPIs in this context serve as tools for assessing the effectiveness of service delivery and the efficiency of resource utilization, rather than profit-driven outcomes.

To address these challenges, additional techniques have been explored to retain the most promising outcomes or even leverage collective input from multiple stakeholders, such as in crowdsourcing \cite{DBLP:conf/mmsys/LoniMGGMAMMVL13,DBLP:conf/socialcom/GalliFMTN12,DBLP:conf/www/BozzonCCFMT12}. These methodologies enable the accurate mapping of processes and ensure comprehensive KPI identification. Other studies have also emphasized integrity checking practices to improve data quality in related applications \cite{CM:LOPSTR2003,CM:FoIKS2004,M:FQAS2004,M:PHD2005,DM:LPAR2006,DM:FlexDBIST2007,DM:FlexDBIST2006,MC:DEXA2005}.
    
\begin{figure*}[!h]
    \centering
    \includegraphics[scale=0.4]{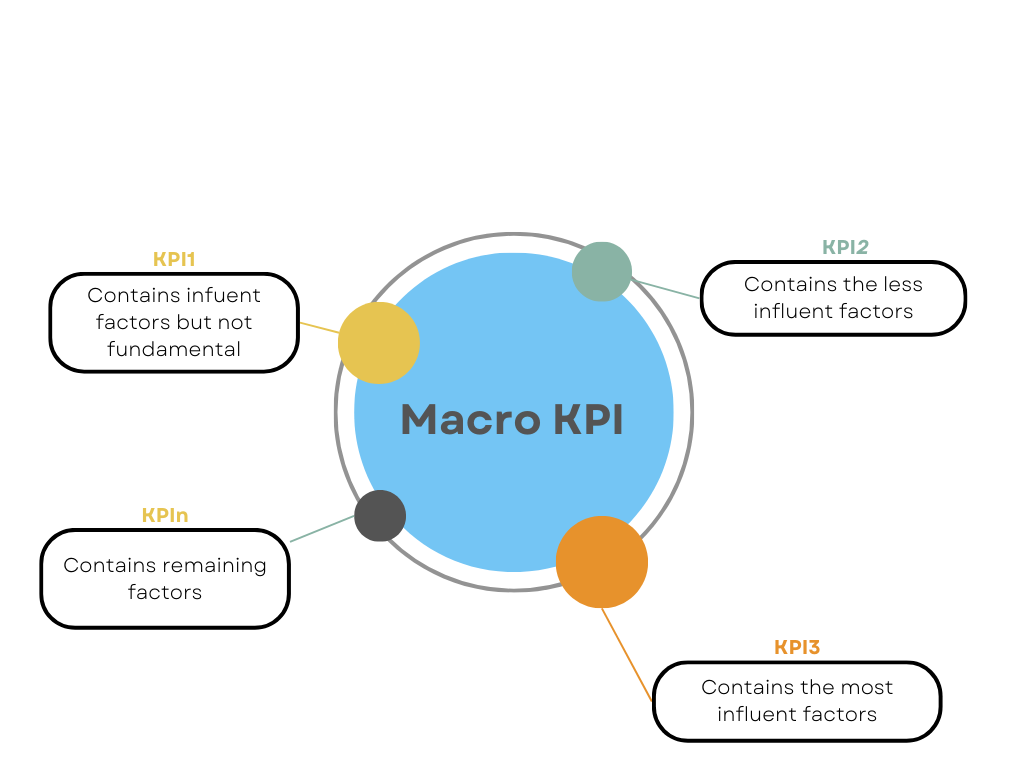}
    \caption{KPI details}
    \label{fig:enter-label}
\end{figure*}

\subsection{Machine Learning to Develop KPIs}

The use of Machine Learning (ML) techniques for developing and optimizing KPIs has gained significant traction across various fields. Several studies demonstrate the versatility of ML in identifying and predicting KPIs suited to specific industry needs.

For example, Ahmed et al. \cite{ahmed2017establishing} employed ML techniques alongside Google Analytics to establish standard rules for identifying optimal KPIs in e-commerce business websites. This study showcases the potential of ML to enhance traditional business tools by offering a structured approach to KPI development.

Fanaei et al. \cite{fanaei2018application} explored the application of ML techniques, such as artificial neural networks (ANN) and neuro-fuzzy methods, to predict overall project KPIs at critical project stages. Their work illustrates the robustness of ML in handling complex datasets commonly found in project management.

Micu et al. \cite{micu2019leveraging} analyzed over a thousand e-commerce websites using ML to identify KPIs linked to company success. Their research highlights the scalability of ML techniques and their utility in extracting actionable insights from large datasets.

El Haddad et al. \cite{el2021machine} investigated the application of ML algorithms for predicting Overall Equipment Effectiveness (OEE) in manufacturing environments. Their findings emphasize the adaptability of ML in improving industrial efficiency through accurate KPI measurement.

Tavakolirad et al. \cite{tavakolirad2023key} introduced an innovative ML-based approach for identifying effective indicators and understanding relationships between them. By integrating supervised and unsupervised models, they analyzed high-risk customers, leveraging clustering algorithms to improve customer segmentation and risk management.

In addition to these examples, we observe that process mining challenges are analogous to issues encountered in query processing under access limitations \cite{CM:ER2008,CCM:JUCS2009}. Other notable research has also investigated using ML in conjunction with process mining logs to enhance efficiency and adaptability \cite{MT:TKDE2011,CM:CIKM2018,CM:TODS2020,MT:PVLDB2010,CCFMT:TODS2013}. These works illustrate the interplay of ML and data-driven methods to refine KPI development across diverse domains.

\section{\uppercase{Framework Proposal}}
\label{fp}
\begin{figure*}[!h]
    \centering
    \includegraphics[scale=0.39]{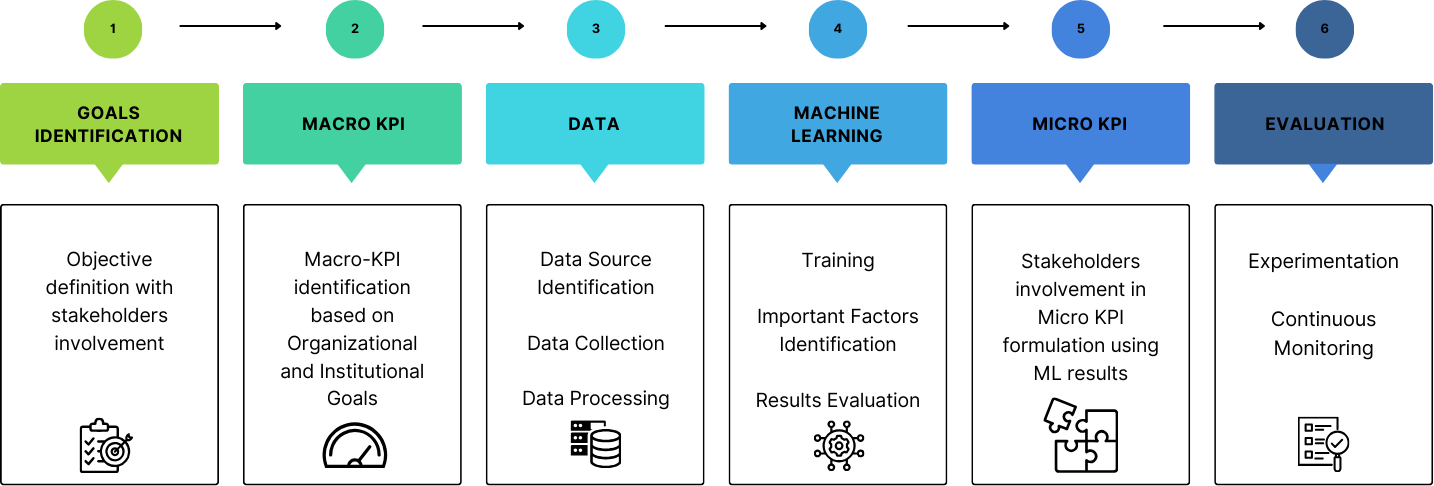}
    \caption{Framework for KPI identification}
    \label{fig:enter-label}
\end{figure*}

In this section we propose a structured framework for performance evaluation in PAs. The proposal involves precise goal-setting, data analysis, and ML techniques. The framework is divided into several key phases, each of them built upon the insights gained from the previous stages. The ultimate goal is to realize an environment supported by stakeholder engagement and continuous improvement.

\begin{enumerate}
        \item \textbf{Goals Identification}: the identification of the objectives of the organization is the initial phase of the framework. In this phase, the organization focuses on the identification and translation of the goals set by superior institutions. Once these goals have been identified, they define the specific organizational objectives. In this phase, it is essential to involve stakeholders to understand their expectations and performance measurement needs. They must be involved for contributing to the goal definition and must be informed about the mission and strategy of the organization. For instance, these objectives may refer to the reduction of response times or increasing citizen satisfaction. 
        \item \textbf{Macro KPI}: based on the goals defined in the previous step, this phase focuses on identifying macro measurements. By identifying macro-KPIs which measure goals achievement, results can be provided for both internal and external purposes. For example, processing time will be considered a macro-KPI for measuring response times. In the justice sector, the time taken to resolve a judgement process can be considered a macro-KPI that measures the goal response times reduction. 
        \item \textbf{Data}: the phase starts with precisely identification of required data, which demands a clear comprehension of the processes to be monitored within the PA. Once the KPIs are established, the next step is the identification of information systems containing the related data. PAs have various data collection systems, such as document archiving databases or human resources management systems, which are vital sources for acquiring the necessary data.
        The next step is the data collection phase, where all pertinent information from the identified systems is extracted. 
        Data processing is the final step before analysis, which involves cleaning, pre-processing, and, if required data enrichment. These preparations are essential to facilitate the effective use of machine learning.
        \item \textbf{Machine Learning}:  
        in this phase it is applied the ML algorithm to the processed data. In particular, RF is effective in handling large volumes of data and identifying the most influential variables with precision.

        It realizes a forest of decision trees, which individually could be subject to over-fitting errors or biased interpretations. However, predictions of many trees are aggregated to obtain a final result which is generally more robust and reliable than single decision tree model result.

        In practical applications, the RF is trained using tabular data that includes input variables, which are specific indicators taken from the event logs of information systems, and a target variable represented by the microscopic KPIs that one wishes to monitor and improve. During the training process, the algorithm analyses the correlation between the input variables and the target, identifying the variables that have the most significant impact on the performance measured by the KPIs.
        
        RF is highly useful in quantifying the importance of each input variable in predicting the KPI. This enables administrators to identify the factors that truly influence results and direct resources towards interventions aimed at improving those aspects. Understanding the variables that play a key role in overall performance is essential for optimizing operations and increasing efficiency.
        
        The information obtained from this process is crucial for developing Micro KPIs.
        
        \item \textbf{Micro KPI}: thanks to the results of the ML algorithm, it is possible to identify the important factors that most contribute to the macro-KPIs identified. These results are then shared with stakeholders which contribute with knowledge domain to the confirmation of the importance of factors. Then, by leveraging the help of stakeholders, these identified factors are unified to make micro KPIs. 
        
        \item \textbf{Experimentation and Evaluation}: the last phase of the framework focuses on testing and evaluation of the proposed solutions. After identifying Micro-KPIs and implementing targeted interventions to improve Macro KPIs, it is crucial to test these changes in real scenarios within the organization.

        During the testing phase, interventions are applied on a small scale or under controlled conditions to monitor their effects and collect meaningful data on the effectiveness of the changes made. 

        The evaluation phase analyses the collected data to determine whether the interventions have led to a concrete improvement in Micro and Macro KPIs. 
        Based on the results obtained, the organisation may decide to extend interventions on a larger scale, make further changes, or possibly discontinue practices that did not bring the desired benefits. This phase is crucial to ensure that operational decisions are evidence-based and to promote continuous improvement within the organisation.
    
\end{enumerate}


\section{\uppercase{Future work and Conclusion}}
\label{conclusion}
This paper presents a framework for constructing Key Performance Indicators in Public Administration scenarios. The framework leverages the RF algorithm to analyze variable importance and identify the most influential factors affecting public service performance. This provides a solid foundation for understanding the critical performance drivers. Additionally, with the integration of knowledge of domain experts, it is possible to develop relevant KPIs. This ensures that our contribution proposal is both theoretically grounded and practically focused. Finally, the resulting KPIs are continuously monitored and adapted, driving PA flexibility in response to changing conditions and ensuring consistent strategies. In addition, the implementation of real-time data analytics would enable instant updates to KPIs, reflecting the dynamic nature of PAs scenarios. 

This work opens up several opportunities for future research. In future work, we plan to explore the application of several ML models to compare their results of the models, and therefore extend our hypothesis. Conducting comparative studies across different PA offices could help to generalize the application of our framework and identify universal best practices. Additionally, tracking the real-world impacts of KPIs adjustments could provide empirical evidence of the benefits of this data-driven approach. Furthermore, aligning public services with community needs could be achieved by prioritising user satisfaction when developing citizen-centring KPIs.

In conclusion, the application of ML techniques, particularly the application of RF and variable importance analysis, represents a step forward towards for a more agile and results-oriented PA. This study extends our understanding of key performance drivers and provides the basis for an effective and targeted performance evaluation system.


\bibliographystyle{unsrt}  
\bibliography{paper}  

\begin{thebibliography}{10}

\bibitem{banu2018measuring}
Geanina~Silviana Banu.
\newblock Measuring innovation using key performance indicators.
\newblock {\em Procedia Manufacturing}, 22:906--911, 2018.

\bibitem{jahangirian2017key}
Mohsen Jahangirian, Simon~JE Taylor, Terry Young, and Stewart Robinson.
\newblock Key performance indicators for successful simulation projects.
\newblock {\em Journal of the Operational Research Society}, 68:747--765, 2017.

\bibitem{kerzner2019using}
Harold Kerzner.
\newblock {\em Using the project management maturity model: strategic planning for project management}.
\newblock John Wiley \& Sons, 2019.

\bibitem{amato2023evolving}
Flora Amato, Simona Fioretto, Eugenio Forgillo, Elio Masciari, Nicola Mazzocca, Sabrina Merola, and Enea~Vincenzo Napolitano.
\newblock Evolving justice sector: An innovative proposal for introducing ai-based techniques in court offices.
\newblock In {\em International Conference on Electronic Government and the Information Systems Perspective}, pages 75--88. Springer, 2023.

\bibitem{dominguez2019taxonomy}
Eladio Dom{\'\i}nguez, Beatriz P{\'e}rez, {\'A}ngel~L Rubio, and Mar{\'\i}a~A Zapata.
\newblock A taxonomy for key performance indicators management.
\newblock {\em Computer Standards \& Interfaces}, 64:24--40, 2019.

\bibitem{velimirovic2011role}
Dragana Velimirovi{\'c}, Milan Velimirovi{\'c}, and Rade Stankovi{\'c}.
\newblock Role and importance of key performance indicators measurement.
\newblock {\em Serbian Journal of Management}, 6(1):63--72, 2011.

\bibitem{parmar2019review}
Aakash Parmar, Rakesh Katariya, and Vatsal Patel.
\newblock A review on random forest: An ensemble classifier.
\newblock In {\em International conference on intelligent data communication technologies and internet of things (ICICI) 2018}, pages 758--763. Springer, 2019.

\bibitem{strobl2008conditional}
Carolin Strobl, Anne-Laure Boulesteix, Thomas Kneib, Thomas Augustin, and Achim Zeileis.
\newblock Conditional variable importance for random forests.
\newblock {\em BMC bioinformatics}, 9:1--11, 2008.

\bibitem{parmenter2015key}
David Parmenter.
\newblock {\em Key performance indicators: developing, implementing, and using winning KPIs}.
\newblock John Wiley \& Sons, 2015.

\bibitem{parmenter2012key}
David Parmenter.
\newblock {\em Key performance indicators for government and non profit agencies: Implementing winning KPIs}.
\newblock John Wiley \& Sons, 2012.

\bibitem{DBLP:conf/mmsys/LoniMGGMAMMVL13}
Babak Loni, Maria Menendez, Mihai Georgescu, Luca Galli, Claudio Massari, Ismail~Seng{\"o}r Alting{\"o}vde, Davide Martinenghi, Mark Melenhorst, Raynor Vliegendhart, and Martha Larson.
\newblock Fashion-focused creative commons social dataset.
\newblock In {\em Multimedia Systems Conference 2013, MMSys '13, Oslo, Norway, February 27 - March 01, 2013}, pages 72--77, 2013.

\bibitem{DBLP:conf/socialcom/GalliFMTN12}
Luca Galli, Piero Fraternali, Davide Martinenghi, Marco Tagliasacchi, and Jasminko Novak.
\newblock A draw-and-guess game to segment images.
\newblock In {\em 2012 International Conference on Privacy, Security, Risk and Trust, PASSAT 2012, and 2012 International Conference on Social Computing, SocialCom 2012, Amsterdam, Netherlands, September 3-5, 2012}, pages 914--917, 2012.

\bibitem{DBLP:conf/www/BozzonCCFMT12}
Alessandro Bozzon, Ilio Catallo, Eleonora Ciceri, Piero Fraternali, Davide Martinenghi, and Marco Tagliasacchi.
\newblock A framework for crowdsourced multimedia processing and querying.
\newblock In {\em Proceedings of the First International Workshop on Crowdsourcing Web Search, Lyon, France, April 17, 2012}, pages 42--47, 2012.

\bibitem{CM:LOPSTR2003}
Henning Christiansen and Davide Martinenghi.
\newblock Simplification of database integrity constraints revisited: A transformational approach.
\newblock In {\em Logic Based Program Synthesis and Transformation, 13th International Symposium LOPSTR 2003, Uppsala, Sweden, August 25-27, 2003, Revised Selected Papers}, volume 3018, pages 178--197. Springer, 2004.

\bibitem{CM:FoIKS2004}
Henning Christiansen and Davide Martinenghi.
\newblock Simplification of integrity constraints for data integration.
\newblock In {\em Foundations of Information and Knowledge Systems, Third International Symposium, FoIKS 2004, Wilhelminenburg Castle, Austria, February 17-20, 2004, Proceedings}, volume 2942, pages 31--48. Springer, 2004.

\bibitem{M:FQAS2004}
Davide Martinenghi.
\newblock Simplification of integrity constraints with aggregates and arithmetic built-ins.
\newblock In {\em Flexible Query Answering Systems, 6th International Conference, FQAS 2004, Lyon, France, June 24-26, 2004, Proceedings}, volume 3055, pages 348--361. Springer, 2004.

\bibitem{M:PHD2005}
Davide Martinenghi.
\newblock {\em Advanced Techniques for Efficient Data Integrity Checking}.
\newblock PhD thesis, Roskilde University, Dept. of Computer Science, Roskilde, Denmark, 2005.

\bibitem{DM:LPAR2006}
Hendrik Decker and Davide Martinenghi.
\newblock A relaxed approach to integrity and inconsistency in databases.
\newblock In {\em Logic for Programming, Artificial Intelligence, and Reasoning, 13th International Conference, LPAR 2006, Phnom Penh, Cambodia, November 13-17, 2006, Proceedings}, volume 4246, pages 287--301. Springer, 2006.

\bibitem{DM:FlexDBIST2007}
Hendrik Decker and Davide Martinenghi.
\newblock Getting rid of straitjackets for flexible integrity checking.
\newblock In {\em Proceedings of the 2nd International Workshop on Flexible Database and Information System Technology (FlexDBIST-07)}, pages 360--364, 2007.

\bibitem{DM:FlexDBIST2006}
Hendrik Decker and Davide Martinenghi.
\newblock Avenues to flexible data integrity checking.
\newblock In {\em Proceedings of the International Workshop on Flexible Database and Information System Technology (FlexDBIST-06) 6 September 2006, Krakow, Poland}, pages 425--429. IEEE Computer Society, 2006.

\bibitem{MC:DEXA2005}
Davide Martinenghi and Henning Christiansen.
\newblock Transaction management with integrity checking.
\newblock In {\em Database and Expert Systems Applications, 16th International Conference, {DEXA} 2005, Copenhagen, Denmark, August 22-26, 2005, Proceedings}, volume 3588, pages 606–--615. Springer, 2005.

\bibitem{ahmed2017establishing}
Haris Ahmed, Tahseen~Ahmed Jilani, Waleej Haider, Mohammad~Asad Abbasi, Shardha Nand, and Saher Kamran.
\newblock Establishing standard rules for choosing best kpis for an e-commerce business based on google analytics and machine learning technique.
\newblock {\em International Journal of Advanced Computer Science and Applications}, 8(5), 2017.

\bibitem{fanaei2018application}
Seyedeh~Sara Fanaei, Osama Moselhi, Sabah~T Alkass, and Zahra Zangenehmadar.
\newblock Application of machine learning in predicting key performance indicators for construction projects.
\newblock {\em methods}, 5(9):1450--1457, 2018.

\bibitem{micu2019leveraging}
Adrian Micu, Marius Geru, Alexandru Capatina, Constantin Avram, Robert Rusu, and Andrei~Alexandru Panait.
\newblock Leveraging e-commerce performance through machine learning algorithms.
\newblock {\em Ann. Dunarea Jos Univ. Galati}, 2:162--171, 2019.

\bibitem{el2021machine}
Choumicha El~Mazgualdi, Tawfik Masrour, Ibtissam El~Hassani, and Abdelmoula Khdoudi.
\newblock Machine learning for kpis prediction: a case study of the overall equipment effectiveness within the automotive industry.
\newblock {\em Soft Computing}, 25(4):2891--2909, 2021.

\bibitem{tavakolirad2023key}
Zahra Tavakolirad, Amir Albadvi, and Elham Akhondzadeh~Noughabi.
\newblock Key performance indicators analysis using machine learning techniques.
\newblock {\em Available at SSRN 4520076}, 2023.

\bibitem{CM:ER2008}
Andrea Cal{\`{i}} and Davide Martinenghi.
\newblock {Conjunctive Query Containment under Access Limitations}.
\newblock In {\em Proceedings of Conceptual Modeling - ER 2008, 27th International Conference on Conceptual Modeling, Barcelona, Spain, October 20-24, 2008}, pages 326--340, 2008.

\bibitem{CCM:JUCS2009}
Andrea Cal{\`{i}}, Diego Calvanese, and Davide Martinenghi.
\newblock {Dynamic Query Optimization under Access Limitations and Dependencies}.
\newblock {\em Journal of Universal Computer Science}, 15(21):33--62, 2009.

\bibitem{MT:TKDE2011}
Davide Martinenghi and Marco Tagliasacchi.
\newblock {Cost-Aware Rank Join with Random and Sorted Access}.
\newblock {\em IEEE Transactions on Knowledge \& Data Engineering}, 24(12):2143--2155, 2012.

\bibitem{CM:CIKM2018}
Paolo Ciaccia and Davide Martinenghi.
\newblock {FA} + {TA} {\textless} fsa: Flexible score aggregation.
\newblock In {\em Proceedings of the 27th {ACM} International Conference on Information and Knowledge Management, {CIKM} 2018, Torino, Italy, October 22-26, 2018}, pages 57--66, 2018.

\bibitem{CM:TODS2020}
Paolo Ciaccia and Davide Martinenghi.
\newblock Flexible skylines: Dominance for arbitrary sets of monotone functions.
\newblock {\em ACM Transactions on Database Systems}, 45(4):18:1--18:45, 2020.

\bibitem{MT:PVLDB2010}
Davide Martinenghi and Marco Tagliasacchi.
\newblock {Proximity Rank Join}.
\newblock {\em Proceedings of the VLDB Endowment}, 3(1):352--363, 2010.

\bibitem{CCFMT:TODS2013}
Ilio Catallo, Eleonora Ciceri, Piero Fraternali, Davide Martinenghi, and Marco Tagliasacchi.
\newblock Top-k diversity queries over bounded regions.
\newblock {\em ACM Transactions on Database Systems}, 38(2):10, 2013.

\end{thebibliography}

\end{document}